# Inference for Belief Networks Using Coupling From the Past

**Michael Harvey**
Dept. of Computer Science
University of Toronto
Toronto, Ontario, Canada
*mikeh@cs.utoronto.ca*

**Radford M. Neal**
Dept. of Statistics and Dept. of Computer Science
University of Toronto
Toronto, Ontario, Canada
*radford@cs.utoronto.ca*

## Abstract

Inference for belief networks using Gibbs sampling produces a distribution for unobserved variables that differs from the correct distribution by a (usually) unknown error, since convergence to the right distribution occurs only asymptotically. The method of "coupling from the past" samples from exactly the correct distribution by (conceptually) running dependent Gibbs sampling simulations from every possible starting state from a time far enough in the past that all runs reach the same state at time $t=0$. Explicitly considering every possible state is intractable for large networks, however. We propose a method for layered noisy-or networks that uses a compact, but often imprecise, summary of a set of states. This method samples from exactly the correct distribution, and requires only about twice the time per step as ordinary Gibbs sampling, but it may require more simulation steps than would be needed if chains were tracked exactly.

## 1    INTRODUCTION

Conditional probabilities for a Bayesian belief network can be calculated exactly only when the network can be transformed into a "junction tree" in which the number of states for each clique is manageable (see Cowell, *et al* 1999). For densely connected networks, clique sizes are large, and the time for exact calculations grows exponentially with the size of the network.

Stochastic simulation methods, in particular Markov chain Monte Carlo methods such as Gibbs sampling, are not limited by the density of edges in the network, but suffer from other problems. A Gibbs sampling simulation converges to the desired invariant distribution asymptotically, but the distribution after a finite time differs from that desired by a generally uncertain amount. This error is greatest in the early part of the chain, which is generally thrown away in a burn-in phase until it is felt that the error has dropped to within the desired tolerance. The burn-in time must usually be estimated because the rate of convergence of the Markov chain is not known theoretically. It is possible to overestimate and waste computing time, or underestimate and include states that are too far from the desired distribution. The conservative user might greatly overestimate the required burn-in time, to minimize the risk of getting the wrong answer.

To overcome this problem of initialization bias, Propp and Wilson (1997) introduced exact sampling, also known as perfect simulation, using the method of "coupling from the past" to obtain states from exactly the desired distribution. Instead of starting a single chain in some arbitrary initial state at time $t = 0$, dependent chains are started in every possible state at some time $T < 0$, far enough back that all the chains coalesce to a single state by time $t = 0$. The state at $t=0$ then comes from the correct distribution, and useful sampling can begin at that time with zero bias (ie, with no systematic error). The final results are still approximate due to ordinary sampling error, but this error is easily controlled.

Propp and Wilson showed how coupling from the past can be implemented efficiently when the state space can be partially ordered, with unique maximal and minimal states, in a way that is preserved through transitions. Belief networks lack this monotonicity property, so some other way of keeping track of all possible chains is needed. We show that it is possible to represent a set of chains by a single summary chain, whose states are sets of states of the original chain. The set of chains may contain both those that are of interest and spurious chains added at intermediate stages of the simulation due to this representation being imperfect. When the summary chain reaches a state that represents a single state of the original chain, coalescence of all the true chains (as well as the spurious chains) will have occurred. The summary chain will then exactly represent the single coalesced chain, whose state at $t = 0$ comes from exactly the desired distribution.

We apply this idea to inference for conditional distributions in layered noisy-or belief networks, in which variables that are siblings are not also directly connected. For these networks, the summary chain transitions can be performed efficiently. The computation time required will depend on the coalescence time of the true chains, plus the overhead caused by any spurious chains. We give examples showing that this overhead can sometimes be large. Other tests on randomly-generated two-layer networks of "diseases" and "symptoms" show that the summary chain method works well enough to be of practical interest. Interestingly, the relative overhead compared to simple Gibbs sampling appears to be smallest for the most difficult problems, for which uncertainty concerning the necessary burn-in time would be greatest.

## 2   NOISY-OR BELIEF NETWORKS

A belief network is a directed acyclic graph in which nodes represent random variables and edges describe how the joint distribution for these variables is expressed. A variable with an edge pointing to another variable is a "parent" of that variable. For each variable $A$ with parents $B_1, \cdots, B_n$, the conditional probabilities $P(A = a \mid B_1 = b_1, \cdots, B_n = b_n)$ are specified. Joint probabilities for all variables, $P(A_1 = a_1, \cdots, A_m = a_m)$, are expressed in terms of the product of all forward conditional probabilities:

$$\prod_i P(A_i = a_i \mid \text{values, } a_j, \text{ for parents, } A_j, \text{ of } A_i)$$

The noisy-OR scheme is a way of specifying these conditional distributions without explicitly listing probabilities for every combination of values for the parent variables. It applies when variables take on values of 0 and 1. Each parent variable influences the child variable to be turned on (value = 1) when it is turned on. The degree of influence is determined by a weight, $c_i$, on the link from parent $X_i$ to child $W$, giving the probability of turning on the child given that the parent $X_i$ is turned on. (If these weights are all one, the scheme is a deterministic OR-gate.) Variable $W$ is also caused to be on for some other reason with probability $p_W$. The conditional probability for $W$ to be on is therefore

$$P(W = 1 \mid \text{values, } x_i, \text{ for parents, } X_i)$$
$$= 1 - (1 - p_W) \prod_{i\,:\,x_i=1} (1 - c_i)$$

## 3   GIBBS SAMPLING

Stochastic simulation using Markov chains was introduced as a way of sampling from conditional distributions for belief networks by Pearl (1987, 1988). The method is now commonly known as Gibbs sampling.

### 3.1   Sampling using Markov chains

A Markov chain is specified by a sequence of discrete random variables $X^{(0)}, X^{(1)}, \cdots$, a marginal distribution, $p_0$, for the initial state $X^{(0)}$, and transition probabilities for state $X^{(t+1)}$ to follow state $X^{(t)}$: $P(x^{(t+1)} \mid x^{(t)})$. The joint distribution of $X^{(0)}, X^{(1)}, \cdots$ is then determined upon making the Markov assumption that $X^{(t)}$ is conditionally independent of $X^{(t-k)}$ for $k > 1$ given $X^{(t-1)}$.

Stationary Markov chains have transition probabilities that do not depend on time, which can be represented with a transition matrix $M$. The value in the $i'th$ row and $j'th$ column of $M$ is the probability of a transition to state $j$ given that the system is in state $i$. The state probabilities at time $t$ (i.e., $P(X^{(t)} = j)$) can be represented as a row vector $p_t$, with $p_t = p_0 M^t$. A distribution $\pi$ is invariant if $\pi = \pi M$. An ergodic Markov chain has an invariant distribution that is reached asymptotically no matter what the initial distribution $p_0$ is, i.e. $\lim_{t \to \infty} p_t = \pi$.

The error in the distribution of the Markov chain at time $t$ can be measured by the total variation distance between the state distribution at that time, $p_t$, and the invariant distribution, $\pi$. For a finite state space $\chi$, this is

$$||p_t - \pi|| = \frac{1}{2} \sum_{x \in \chi} |p_t(x) - \pi(x)|.$$

Asymptotically, the error decays exponentially as

$$error = ae^{-t/c}$$

where $a$ and $c$ are constants specific to the Markov chain. (See Rosenthal (1995) for further discussion.) If the user has an error tolerance $\epsilon$, the burn-in time for Gibbs sampling should be

$$burn\ in\ time = -c \ln(\epsilon/a)$$

However, the convergence behaviour of the Markov chain is usually unknown (i.e., the constants $a$ and $c$ are not known). The conservative user will try to choose as large a burn-in time as is practical to lessen the chances of obtaining erroneous results.

### 3.2   Gibbs sampling for belief networks

When the state, $X$, consists of several variables, $X_1, \ldots, X_n$, a transition matrix that leaves a desired distribution $\pi$ invariant can be built from a sequence of matrices representing transitions that change only a single variable. The transition matrix $M$, on the state space of all possible combinations of values for $X_1, \ldots, X_n$, is written as $M = B_1 B_2 \cdots B_n$, with $B_k$ representing an update that changes only $X_k$ (i.e., entries in $B_k$ for transitions that change variables other than $X_k$ are zero). In the Gibbs sampling scheme, the entries in $B_k$ corresponding to changes in $X_k$ alone are the conditional probabilities under $\pi$ of the variable $X_k$ taking on its various values, given the current values of the other variables. Each such $B_k$ will leave the distribution $\pi$ invariant, and hence so will $M$.

Pearl (1987) derives Gibbs sampling transition probabilities for a belief network. When other variables are fixed, conditional probabilities for a particular variable depend only on its parents, its children, and its children's parents. From the definition of belief networks,

$$P(V_k|V_1, , \cdots, V_{k-1}, V_{k+1}, \cdots, V_n)$$
$$\propto P(V_k|V_1, \cdots, V_{k-1}) \prod_{j>k} P(V_j|V_1, \cdots, V_k, \cdots, V_{j-1})$$

Here $V_1, V_2, \cdots, V_{k-1}$ are possible parents of $V_k$, and $V_j$ for $j > k$ are possible children of $V_k$, since $V$ is ordered with parents before children. This expression can be evaluated using the specified conditional distributions for a variable given its parents. When $V_j$ is not actually a child of $V_k$, the corresponding factor can be omitted.

The effect of sampling from the distribution of unknown variables conditional on known values for other variables is achieved by simply keeping the values of the known variables fixed while the others are updated by Gibbs sampling, using the above probabilities.

For noisy-or belief networks, if a child of the variable being updated has the value 0, the other parents of that child can be ignored in the calculation — child variables instantiated to 0 cannot transmit information between parents.

## 4 EXACT SAMPLING

Propp and Wilson (1997) proposed exact sampling using coupling from the past as a way to eliminate the error from using finite-length simulation runs. An ergodic Markov chain will reach its equilibrium distribution if it is run for an infinite amount of time. Therefore, if one were willing to wait forever, one could be sure that the correct distribution of the Markov chain had been reached. Propp and Wilson show that it is not necessary to wait forever to arrive at this result, however — that, at least when the state space is finite, there is a way to find the exact result with a finite number of computations. The state found in this way may be used as the starting state for a Gibbs sampling run that will be free of bias.

### 4.1 The idea of coupling from the past

Propp and Wilson's idea is to run many chains from some time, $T < 0$, in the past, starting from every possible state. These chains are "coupled", by introducing dependencies between their transitions, in an attempt to make them coalesce to the same state by $t = 0$. If by time $t = 0$ they have all coalesced into one chain, then it can be said that no matter what state was started from at time $t = T$, the same state at time $t = 0$ results. If coalescence of chains started at $t = T$ does not occur by time $t = 0$, then the procedure is repeated from further back in the past.

At a minimum, the chains started from the various initial states must be dependent to the extent that two chains arriving at the same state henceforth use the same pseudo-random numbers for all their subsequent transitions. This causes chains to stay together once they first coalesce. Coalescence of all chains to a single state can be encouraged by introducing dependencies beyond this. Such dependencies between chains do not invalidate the results as long as transitions made at different times remain independent. In this paper, we will consider only systematic Gibbs sampling, in which the variables of the network are updated in some sequence. Randomness is therefore required only for setting the value of the variable being updated according to its conditional distribution. A single real-valued pseudo-random number is sufficient for making this random choice at each time step. Dependencies between chains are introduced by using the same such random number for all chains. If the chains are restarted from further back in the past, the same pseudo-random numbers as before are used at those times that were previously visited.

Propp and Wilson show that if the chain is ergodic, this coupling from the past procedure will, with probability 1, lead to coalescence at $t = 0$ once the chains are started from sufficiently far in the past, and that the unique state of the coalesced chains at $t = 0$ is distributed exactly according to the chain's equilibrium distribution.

When coalescence does not occur, it would be inefficient to try again with a chain started just one time step further back, since the new chain must be run all the way to time $t = 0$. It is more efficient to start runs at times $t = -1, -2, -4, -8, -16, \cdots$ until coalescence finally occurs. Propp and Wilson (1997) show that this scheme is not far from optimal, requiring no more than four times the total number of simulation steps as would be needed if the actual coalescence time were somehow guessed.

Note that a coalesced chain must be continued to time $t = 0$ to obtain a state from the correct distribution. Usually, coalescence occurs before reaching time $t = 0$, but if the state at that time is used, there is a bias introduced toward conditions that favour coalescence of chains. By always selecting the state at time $t = 0$, there is no dependence on the time that coalescence occurs. Also note that when the chains do not all coalesce, it is not valid to just throw away these chains and start new chains from further back. Instead, the old chains must be extended backwards in time — i.e., the old pseudo-random numbers generated at times that were already visited must be re-used. Otherwise, a bias is introduced by a preference for pseudo-random numbers that more easily allow coalescence.

### 4.2 Using states obtained by exact sampling

The coupling from the past procedure just described can be performed a number of times, each time with new pseudo-random variables, thus obtaining multiple independent states from exactly the desired distribution. Each time the procedure is run, it must search for a starting time that allows the chains to coalesce. The procedure typically requires varying amounts of running time to complete (i.e., one must go back varying amounts of time in the past to cause the chains to coalesce). In practice, if any of these independent attempts result in coalescence, it is reasonable to expect all such attempts to succeed within a reasonable time period, since the chances of coalescing further back in the past are independent of the failure to coalesce later on. However, if none of the runs coalesce in a reasonable time, one must declare the results indeterminate, which is superior to getting a wrong answer using Gibbs sampling with the same number of steps.

A state from the invariant distribution found using coupling from the past may be used to initialize an ordinary Gibbs sampling run, which continues forward from $t = 0$. The states at times $t \geq 0$ will all have exactly the correct distribution. However, it is desirable to run the coupling from the past procedure several times, in order to find a number of initial states from the invariant distribution, and to take samples from each of the chains that follow them. These initial states are more valuable than the states that follow them, since they are completely independent of each other, but they come at the cost of coupling chains from the past. At the same time, the following states are less valuable because of their dependence on prior states, but they can be produced at the much lower cost of one Markov chain transition.

### 4.3 Efficiently tracking chains

Keeping track of every chain for every possible starting state is generally infeasible, since the size of the state space is exponential in the number of variables in the network. Propp and Wilson (1997) show that coupling from the past can be implemented efficiently when states can be given a partial order that is preserved through Markov chain transitions, by simulating just two chains, started from the minimal and maximal states. Furthermore, they show that for such monotonic chains, coalescence cannot be much slower than convergence of the chain. This makes coupling from the past quite attractive for such problems.

Belief network states usually cannot be ordered in a way that makes the Markov chain monotonic. An alternative way of keeping track of every chain is therefore needed. There is also no known guarantee that the coalescence time will not be much greater than the time required for Gibbs sampling to converge to close to the desired distribution, though we know of no examples of this occurring.

To address the efficiency issue, a scheme is needed to simplify the tracking of all the chains. For noisy-or belief networks, we attempt to summarize the chains with one chain whose states are sets of states of the original chain. The amount of work for each transition of the summary chain is the same as for two transitions of the original chain, provided that no sibling variables in the network are directly connected. However, our method does not track the set of chains precisely, which may slow detection of coalescence, though the final result is still from the exactly correct distribution. Similar techniques, applied to simulation of Markov random fields, have been independently developed by Huber (1998) and by Häggström and Nelander (1999).

## 5   EXACT SAMPLING FOR NOISY-OR BELIEF NETWORKS

We now show how coalescence of a large number of Markov chains for a layered noisy-or belief network can be determined by simulating one summary chain, whose states approximate sets of states of chains started in all possible initial states.

### 5.1 Approximating a set of states

The state space $S$ of noisy-or belief networks has variables that take the values 0 or 1. We approximate a set of states in $S$ by a single state in a state space $S^{(?)}$, in which variables take the values 0, 1, or ?. The mapping, $\beta$, from such a state to a set of states of the

original space is

$$\beta(V^{(?)}) = \{ V \in S : \text{for all } i, V_i^{(?)} = V_i \text{ or } V_i^{(?)} = ? \}$$

That is, $\beta$ selects all the states in $S$ where every variable matches the corresponding variable in $S^{(?)}$, with ? matching either 0 or 1. For example,

$$\beta(1?001?) = \left\{ \begin{array}{c} 100010 \\ 100011 \\ 110010 \\ 110011 \end{array} \right\}$$

Not every set of states in $S$ can be represented exactly by a state in $S^{(?)}$. For instance, there is no exact representation of

$$\left\{ \begin{array}{c} 100010 \\ 100011 \\ 110010 \end{array} \right\}$$

Hence, to avoid losing true states, the approximation will sometimes have to include spurious states as well.

### 5.2 Approximating a set of chains

We now show how a single summary chain on $S^{(?)}$ can be simulated so as to approximate a set of chains on $S$, with none of the true chains being lost, though spurious chains may be introduced, delaying coalescence.

Like the original Gibbs sampling chain, transitions in the summary chain change one variable at a time. For each state in $S$, the variable being updated has some conditional probability of changing to a 1, given the other variables in that state. Over the set of states that the state of the summary chain maps to, this conditional probability will have some maximum and some minimum value, which can be used to determine the transition probabilities of the variable in $S^{(?)}$, using the fact that the chains are coupled by using the same pseudo-random variable, $U$. We that assume $U$ is uniformly distributed over $[0, 1))$, and that transitions in the original chain are determined by setting the variable being updated to 1 if $U$ is less than the conditional probability of a 1, and to 0 otherwise. Transitions in the summary chain can then be determined as follows:

- If $U$ is less than the minimum probability, then all the original chains would set the variable to 1. Set the variable in the summary chain to 1 also.

- Similarly, if $U$ is greater then or equal to the maximum probability, set the variable in the summary chain to 0.

- If $U$ is between the minimum and maximum probabilities, then some of the original chains would set the variable to 0 and some would set it to 1. To represent this, set the variable in the summary chain to ?.

Transitions done this way do not lose track of any of the true chains, but spurious chains may be introduced. For example, consider a transition in the summary chain that changes the last variable of 1?0010 (which $\beta$ maps to $\{100010, 110010\}$), and suppose that in the original chains, the transitions from these two states are to states with different values for the last variable (e.g., 100010 $\to$ 100010 and 110010 $\to$ 110011). The transition in the summary chain will have to be to the state 1?001?, which maps to four states of $S$, rather than the previous two states. Hence two spurious chains have been introduced.

### 5.3 Efficient simulation of the summary chain

It is possible to determine the minimum and maximum of the conditional probability used in updating $V_k$, which is $P(V_k = 1 \mid V_1, \ldots, V_{k-1}, V_{k+1}, \ldots, V_n)$, over all $V$ in $\beta(S^{(?)})$, without exhaustive search, by examining just two judiciously chosen two states in $\beta(S^{(?)})$, for which this conditional probability will take on its minimum and maximum value. Let $pa(V_i)$ be parents of $V_i$, and $c(V_i)$ be the children of $V_i$. The required conditional probability is minimized or maximized by minimizing or maximizing the ratio

$$\frac{P(V_k = 1 \mid pa(V_k)) \times \prod_{V_j \in c(V_k)} P(V_j \mid pa(V_j), V_k = 1)}{P(V_k = 0 \mid pa(V_k)) \times \prod_{V_j \in c(V_k)} P(V_j \mid pa(V_j), V_k = 0)}$$

The rules for selecting the appropriate states are summarized below, and justified in detail by Harvey (1999).

For the minimum probability that $V_k = 1$, look at the $V \in \beta(V^{(?)})$ for which:

> If $V_j$ is a child or parent of $V_k$ and $V_j^{(?)} = ?$, then $V_j = 0$.

> If $V_p$ is a parent of some child, $V_j$, of $V_k$, and $V_j^{(?)} = 1$, and $V_p^{(?)} = ?$, then $V_p = 1$.

For the maximum probability that $V_k = 1$, look at the $V \in \beta(V^{(?)})$ for which:

> If $V_j$ is a child or parent of $V_k$ and $V_j^{(?)} = ?$, then $V_j = 1$.

> If $V_p$ is a parent of some child, $V_j$, of $V_k$, and $V_j^{(?)} \in \{1, ?\}$, and $V_p^{(?)} = ?$, then $V_p = 0$.

Note that variables not mentioned above are irrelevant to computing the conditional probability for $V_k$.

These rules assume that sibling variables are not directly connected. That is, a child does not share a parent with its parent $V_k$. If this were so, the two possible settings for the value of that parent could have opposite effects on the conditional probability ratios for $V_k$ being 0 or 1 given its parents, and for the child of $V_k$ having its present value given that $V_k$ is 0 or 1. For layered networks, in which edges go only from variables in one layer to those in the layer immediately below, this problem does not arise.

### 5.4 Time required

The summary chain is started at some time $T < 0$ in the state where all variables are set to ?, representing the set of all possible states. If none of the states of the network have zero probability, it is not hard to show that if $T$ is early enough, then by time $t = 0$, the summary chain will have reached a state representing a single network state — i.e., one in which none of the variables are set to ?. Since no chains are lost in the approximation used, the true set of chains will have coalesced when this happens. However, because of spurious chains, the summary chain may need to be started at an earlier time than would be needed if all chains were tracked explicitly.

The *coalescence time* for a simulation is the minimum time in the past sufficient to produce coalescence by $t = 0$. If simulation is done using start times of $-T = 1, 2, 4, 8, ...$, Propp and Wilson (1997) show that the expected total number of time steps simulated is around 2.89 times the coalescence time. Our noisy-or simulation scheme requires two calculations of conditional probability at each time step, in order to obtain the minimum and the maximum. Therefore, the expected computational work, measured in terms of computations similar to ordinary Gibbs sampling updates, is 5.78 times the expected coalescence time.

## 6 COALESCENCE TIMES

When the summary chain method is used, the coalescence time may be greater than if chains were tracked explicitly. Here we look at the relationship between these two coalescence times for some small problems by calculating the convergence rates of the original and summary chains from the eigenvalues of their transition matrices. The convergence rate of the original chain can be bounded in terms of its expected coalescence time, and though the reverse is not true in general, we will here take the convergence rate to be indicative of the chain's expected coalescence time.

### 6.1 Transition matrix eigenvalues

The convergence rate of an ergodic Markov chain is related to the magnitudes of the eigenvalues of the transition matrix $M$, as reviewed, for example, by Rosenthal (1995). There is a (left) eigenvector corresponding to the invariant distribution, $\pi$, of the Markov chain, with eigenvalue equal to 1. The magnitudes of the other (possibly complex) eigenvalues are less than 1.

If the Markov chain is started in the initial distribution $p_0$, the distribution at time $n$ will be $p_n = p_0 M^n$. As $n \to \infty$, $p_n \to \pi$ for an ergodic chain, with the rate of convergence being determined by the magnitude of the eigenvalue whose magnitude is second-largest (i.e., the largest other than the eigenvalue associated with $\pi$, whose value is 1). This eigenvalue will have a magnitude less than one, but the closer it is to one, the slower will be the convergence of the Markov chain.

The transition matrix of a summary chain will have all the eigenvectors and eigenvalues of the original chain, including the eigenvector and eigenvalue corresponding to the invariant distribution, to which both chains converge (if the original chain is ergodic). The summary chain will also have some eigenvectors and eigenvalues associated with states where some variables have ? as their value, and some of these eigenvalues may be larger in magnitude than the second largest eigenvalue of the original chain. Therefore, the summary chain cannot converge any faster than the chains it summarizes, and may converge more slowly.

Below, we examine these eigenvectors and eigenvalues for some simple diagnostic networks in which a top layer of variables represent "diseases", which can cause various "symptoms", represented by variables in the bottom layer. Interest focuses on inference for what diseases are present, given certain observed symptoms.

### 6.2 Perfectly summarized networks

The transitions of the chains being coupled from the past can be perfectly summarized in certain types of networks. A trivial example is a one-disease network with one or more symptoms that are known. When there is just one ? variable in the summary state, it is an exact representation of a set of two states.

Any network with two unknowns can also be perfectly summarized. Although it is possible that the state of the summary chain does not exactly represent the set of states of the true chains, each variable of the summary chain summarizes the set of values for that variable correctly, even if some constraints between variables are missed. (This can be shown by induction, starting with the initial situation where all values are possible, and all summary variables are set to ?.) When the true chains all coalesce, the summary chain will therefore show coalescence as well (i.e., it will have no ? values).

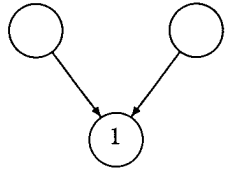

*Disease apriori probability:* 0.1
*Symptom apriori probability:* 0.0
*Noisy-or weight:* 1.0

Figure 1: A network with two diseases.

*Eigenvalues and eigenvectors of the original chain*

| states | 1 | .81 |
|---|---|---|
| 00 | 0 | 0 |
| 01 | 0.47368 | 0.49809 |
| 10 | 0.47368 | −0.44828 |
| 11 | 0.05263 | −0.04981 |

*Eigenvalues and eigenvectors of the summary chain*

| states | 1 | .81 | .81 |
|---|---|---|---|
| 00 | 0 | 0 | 0 |
| 0? | 0 | 0 | 0 |
| 01 | 0.47368 | 0.49809 | −0.44178 |
| ?0 | 0 | 0 | 0 |
| ?? | 0 | 0 | 0.49391 |
| ?1 | 0 | 0 | 0.05488 |
| 10 | 0.47368 | −0.44828 | −0.09631 |
| 1? | 0 | 0 | 0 |
| 11 | 0.05263 | −0.04981 | −0.01070 |

Table 1: Eigenvectors and eigenvalues of transition matrices for the two-disease network.

An example two-disease network is shown in Figure 1. The original Markov chain has $2^2 = 4$ states for the two disease variables (00, 01, 10, 11). The summary chain has $3^2 = 9$ states (00, 0?, 01, ?0, ??, ?1, 10, 1?, 11). The non-zero eigenvalues of the transition matrices, with associated eigenvectors (as column vectors) are shown in Table 1.

The largest eigenvalue in the summary chain apart from the eigenvalue of 1 is 0.81, the same as the second-largest eigenvalue of the original chain. This suggests that coalescence using the summary chain will be just as quick as when explicitly tracking chains from all possible initial states, and indeed experimental results show that both methods perform identically.

Simple networks such as these are not of great interest by themselves, but they may exist as sub-networks within a larger network. In a noisy-or network of symptoms and diseases, a symptom that is known to be absent does not produce interactions between the diseases that are its potential causes. Sections of the network can effectively be independent, with sub-states whose transitions do not interact with those of other sections. If a large network consists only of independent single-disease sub-networks, for example, exact sampling will converge in one iteration. Such situations are discovered at run-time, depending on how the symptoms are instantiated.

### 6.3 Imperfectly summarized networks

In general, network transitions cannot be perfectly summarized by a chain in which states consist of 0/1/? variables, and this can lead to worse convergence for such a summary chain than for the original chain. A moderate example of this is shown in Figure 2.

The non-zero eigenvalues and associated eigenvectors of the summary chain transition matrix for this network are shown in Table 2. Some of these are the same as for the transition matrix of the original chain. The additional eigenvector, which has non-zero components for states with ?-valued variables, has an eigenvalue whose magnitude is greater than that of all the other eigenvalues less than 1. Its magnitude of .97, compared to the next largest, $\sqrt{0.851^2 + 0.075^2} \approx 0.85$, indicates moderately worse convergence for the summary chain. Experiments with this network show that detecting coalescence by $t = 0$ using the summary chain requires starting on average 53.9 time steps in the past, compared to only 17.6 time steps if every state is tracked explicitly.

If the probabilities in the imperfectly summarized network are more extreme, the summary chain is much worse than tracking every state. Figure 3 shows a network that coalesces quickly when every state is tracked explicitly, and hence must also converge quickly when ordinary Gibbs sampling is done, but for which a very long time is needed (on average) for coalescence to be detected when using the summary chain. This summary chain's second-largest eigenvalue is .996, compared to 0.352 for the original chain. The non-zero apriori symptom probability in this network helps the set of chains to coalesce quickly by allowing for other explanations of the evidence besides the diseases of the network — an effect that is lost when sets of states are approximately summarized.

## 7 EMPIRICAL TESTS

Although the above results show that coupling from the past using a summary chain can perform poorly, empirical testing shows that it performs reasonably well on simulated two-level diagnostic networks. These

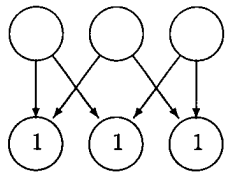

*Disease apriori probability:* 0.1
*Symptom apriori probability:* 0.0
*Noisy-or weight:* 1.0

Figure 2: An imperfectly-summarized network.

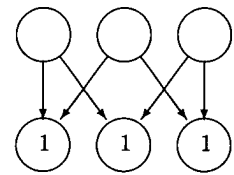

*Disease apriori probability:* 0.001
*Symptom apriori probability:* 0.001
*Noisy-or weight:* 1.000

Figure 3: Extreme imperfectly-summarized network

| states | 1 | 0.973 | $0.851 + 0.075i$ | $0.851 - 0.075i$ |
|---|---|---|---|---|
| 000 | 0 | 0 | 0 | 0 |
| 00? | 0 | 0 | 0 | 0 |
| 001 | 0 | 0 | 0 | 0 |
| 0?0 | 0 | 0 | 0 | 0 |
| 0?? | 0 | 0 | 0 | 0 |
| 0?1 | 0 | 0 | 0 | 0 |
| 010 | 0 | 0 | 0 | 0 |
| 01? | 0 | 0 | 0 | 0 |
| 011 | 0.321 | $-0.193$ | $0.338 + 0i$ | $0.338 + 0i$ |
| ?00 | 0 | 0 | 0 | 0 |
| ?0? | 0 | 0 | 0 | 0 |
| ?01 | 0 | 0 | 0 | 0 |
| ??0 | 0 | 0 | 0 | 0 |
| ??? | 0 | 0.441 | 0 | 0 |
| ??1 | 0 | 0.049 | 0 | 0 |
| ?10 | 0 | 0 | 0 | 0 |
| ?1? | 0 | 0.049 | 0 | 0 |
| ?11 | 0 | 0.005 | 0 | 0 |
| 100 | 0 | 0 | 0 | 0 |
| 10? | 0 | 0 | 0 | 0 |
| 101 | 0.321 | $-0.191$ | $-0.152 - 0.283i$ | $-0.152 + 0.283i$ |
| 1?0 | 0 | 0 | 0 | 0 |
| 1?? | 0 | 0.045 | 0 | 0 |
| 1?1 | 0 | 0.005 | 0 | 0 |
| 110 | 0.321 | $-0.189$ | $-0.167 + 0.254i$ | $-0.167 - 0.254i$ |
| 11? | 0 | 0 | 0 | 0 |
| 111 | 0.036 | $-0.021$ | $-0.019 + 0.028i$ | $-0.019 - 0.028i$ |

Table 2: Eigenvectors and eigenvalues of the summary chain for the imperfectly-summarized network.

tests are described in detail by Harvey (1999).

Tests were first performed using randomly-generated networks with 10 possible diseases and 10 symptoms whose structure was thought to be plausible for an actual application. The performance of coupling from the past using the summary chain was then compared to performance when every chain is tracked explicitly. For these problems, coalescence of the summary chain took at most 512 iterations, and using the summary chain was never hugely worse than tracking all chains.

Further tests were run on networks of the same size that were constructed randomly so as to produce a larger fraction of difficult problems. In these tests, the summary chain tended to work well on the problems for which coalescence was slowest (even tracking every chain explicitly), with coalescence typically detected using the summary chain in only about twice the time that was required when tracking every chain explicitly.

When coalescence is slow even when tracking all states explicitly, it is likely (though not guaranteed) that Gibbs sampling would also converge slowly. Nevertheless, calculations show that if the convergence properties of the chain for these problems were somehow known, use of coupling from the past would be computationally favourable only for users with quite a low error tolerance. In practice, the convergence properties are not known, however, which makes relying on a "burn-in" period problematic. Coupling from the past eliminates all uncertainty about whether the right distribution has been reached, which may make it attractive whenever it is computationally feasible.

## References

Cowell, R. G., Dawid, A. P., Lauritzen, S. L., and Spiegelhalter, D. J. (1999) *Probabilistic Networks and Expert Systems*, Springer-Verlag.

Häggström, O. and Nelander, K. (1999) "On exact simulation of Markov random fields using coupling from the past", *Scandinavian Journal of Statistics*, vol. 26, pp. 395-411.

Harvey, M. (1999) *Monte Carlo Inference for Belief Networks Using Coupling From the Past*, MSc Thesis, Computer Science, University of Toronto.

Huber, M. (1998) "Exact sampling and approximate counting techniques", *Proceedings of the 30th ACM Symposium on the Theory of Computing*, pp. 31-40.

Propp, J. G. and Wilson, D. B. (1996) "Exact Sampling with Coupled Markov Chains and Applications to Statistical Mechanics", *Random Structures and Algorithms*, vol. 9, pp. 223-252.

Pearl, J. (1987) "Evidential Reasoning Using Stochastic Simulation of Causal Models", *Artificial Intelligence*, vol. 32, pp. 245-257.

Pearl, J. (1988) *Probabilistic Reasoning in Intelligent Systems: Networks of Plausible Inference*, Morgan Kaufmann.

Rosenthal, J. S. (1995) "Convergence rates of Markov chains", *SIAM Review*, vol. 37, pp. 387-405.